\documentclass[letterpaper, 10 pt, conference]{ieeeconf}  % Comment this line out if you need a4paper

\IEEEoverridecommandlockouts                              % This command is only needed if 
\usepackage{graphicx}
\usepackage{adjustbox}
\usepackage{footmisc}

\usepackage{amsfonts} 
\usepackage[ruled,vlined]{algorithm2e}
\usepackage{amsmath}
\usepackage{hyperref}
\usepackage{booktabs}

\title{\LARGE \bf Understanding Dynamic Scenes using Graph Convolution Networks }
\author{ Sravan Mylavarapu*\textsuperscript{1}\thanks{$^{1}$Authors are with the Center for Visual Information Technology, KCIS, IIIT Hyderabad}, Mahtab Sandhu*\textsuperscript{2}\thanks{$^{2}$Authors are with the Robotics Research Center, KCIS, IIIT Hyderabad}, Priyesh Vijayan\textsuperscript{3}\thanks{$^{3}$ School of Computer Science, McGill University and Mila},\\ K Madhava Krishna\textsuperscript{2} , Balaraman Ravindran\textsuperscript{4}\thanks{$^{4}$ Dept. of CSE and Robert Bosch Center for Data Science and AI, IIT
Madras}, Anoop Namboodiri\textsuperscript{1}}

\begin{document}

% \makeatletter
% \let\@oldmaketitle\@maketitle
% \renewcommand{\@maketitle}{\@oldmaketitle
% \centering
% % \includegraphics[width=17.5cm]{images/Teaser.pdf}
% \includegraphics[trim=0 0 0 0,clip,  width=0.85\linewidth]{images/Teaser.pdf}
% % \vspace{-0.1cm}
% \captionof{figure}{Fig.}
% \label{fig:teaser}
% }
% \makeatother

\maketitle
\thispagestyle{empty}
\pagestyle{empty}

%%%%%%%%%%%%%%%%%%%%%%%%%%%%%%%%%%%%%%%%%%%%%%%%%%%%%%%%%%%%%%%%%%%%%%%%%%%%%%%%
\begin{abstract}
We present a novel Multi-Relational Graph Convolutional Network (MRGCN) based framework to model on-road vehicle behaviors from a sequence of temporally ordered frames as grabbed by a moving monocular camera. The input to MRGCN is a multi-relational graph where the graph's nodes represent the active and passive agents/objects in the scene, and the bidirectional edges that connect every pair of nodes are encodings of their Spatio-temporal relations.

We show that this proposed explicit encoding and usage of an intermediate spatio-temporal interaction graph to be well suited for our tasks over learning end-end directly on a set of temporally ordered spatial relations. We also propose an attention mechanism for MRGCNs that conditioned on the scene dynamically scores the importance of information from different interaction types.

The proposed framework achieves significant performance gain over prior methods on vehicle-behavior classification tasks on four datasets. We also show a seamless transfer of learning to multiple datasets without resorting to fine-tuning. Such behavior prediction methods find immediate relevance in a variety of navigation tasks such as behavior planning, state estimation, and applications relating to the detection of traffic violations over videos.

% We present a novel Multi Relational Graph Convolutional Network (MRGCN) based framework to model on-road vehicle behaviours from a sequence of temporally ordered frames as grabbed by a moving monocular camera. The input to MRGCN is a Multi Relational Graph (MRG) where the nodes of the graph represent the active and passive participants/agents in the scene while the bidrectional edges that connect every pair of nodes are encodings of the spatio-temporal relations. The bidirectional edges of the graph encode the  temporal interactions between the agents that constitute the two nodes of the edge. The proposed method of obtaining this encoding is shown to be specifically suited for the problem at hand as it outperforms more complex end to end learning methods that do not use such intermediate representations of evolved spatio-temporal relations between agent pairs. We show significant performance gain in the form of behaviour classification accuracy on a variety of datasets from different parts of the globe over prior methods as well as show seamless transfer without any resort to fine tuning across multiple datasets. Such behaviour prediction methods find immediate relevance in a variety of navigation tasks such as behaviour planning, state estimation as well as in applications relating to detection of traffic violations over videos.

\end{abstract}

%%%%%%%%%%%%%%%%%%%%%%%%%%%%%%%%%%%%%%%%%%%%%%%%%%%%%%%%%%%%%%%%%%%%%%%%%%%%%%%%
\section{INTRODUCTION}

\begin{figure}[!t]
    \centering
    \includegraphics[trim=0 0 0 0,clip,  width=0.85\linewidth]{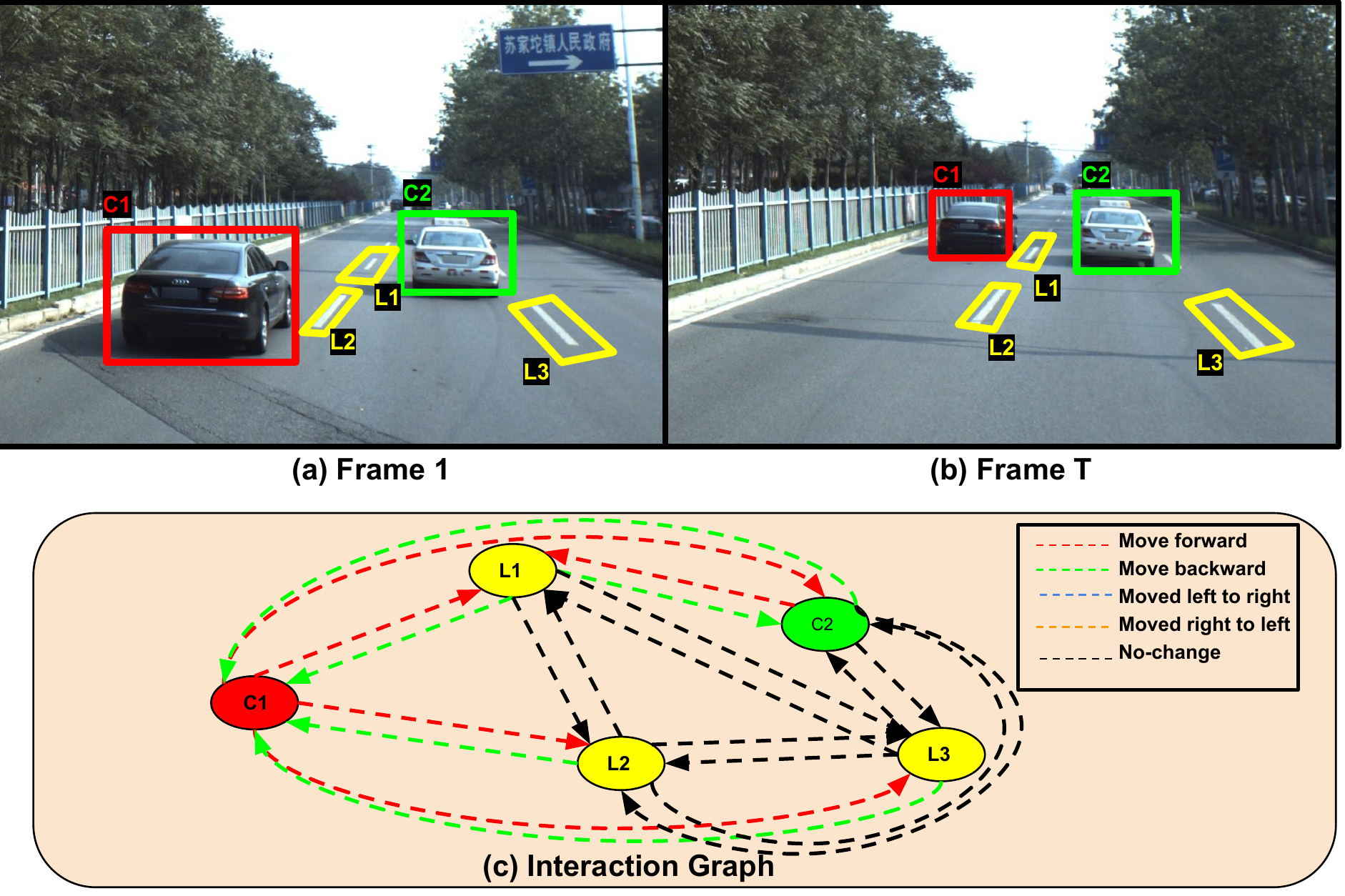}
    \caption{\scriptsize 
    The figures (a) and (b)  show two cars and three lane-markings from two different time frames.  The evolution of the whole scene is captured in the Interaction graph (c). Our proposed model infers over such Interaction graphs to classify objects' behaviors. Here, the car on the left is moving ahead of lane markings and the car on the right. With the Interaction graph (c), our model can predict that the car on the left is overtaking the right-side car.}
    \vspace{-6mm}
    \label{fig:teaser}
\end{figure}

We consider dynamic traffic scenes consisting of potentially active participants/agents such as cars and other vehicles that constitute the traffic and passive objects such as lane markings and poles (see example in Fig. \ref{fig:teaser}). In this work, we propose a framework to model the behavior of each such active agents by analyzing the Spatio-temporal evolution of their relations with other active and passive objects in the scene. By relation, we refer to the spatial relations an agent/object possesses and enjoys with other agents/objects, such as between the vehicle and lane markings, as shown in Fig. \ref{fig:teaser}(c). 

% \textcolor{blue}{Interpreting traffic scenes in terms of constituent vehicle behaviors happens naturally for humans. Indeed most often this awareness, for example, a realization that a ``car is cutting into my lane from behind me,” results in intelligent driving decisions despite lack of intricate scene understanding in terms of vehicle poses, trajectories, and coordinates}. 

Here, we model both objects and agents, and thus for convenience, we commonly refer them as objects and specifically as agents when referring to active vehicles. The evolution of the spatial relationship between all pairs of objects in a scene is essential in understanding their behaviors. To this end, we propose an Interaction graph that models different agents and objects in the scene as nodes. This graph captures the Spatio-temporal evolution of relations between all-pair of objects in the scene with appropriate bi-directional asymmetric edges with annotations reflecting their evolution (see Fig. \ref{fig:teaser}(c)). 

The dynamic traffic scene modeled as an Interaction graph is then inputted to a Multi-Relational Graph Convolutional Network (MRGCN), which outputs the overall behavior exhibited by all the agents and objects in the scene. While the MRGCN maps the input graph to an output behavior for every graph node, we are only interested in active and not the passive objects. Hitherto by behavior, we denote the overall behavior of an active agent in the scene.  For example, in Figure \ref{fig:teaser}, the behavior of the car on the left is \textit{Overtaking}, and the car on the right is \textit{moving ahead}.

The choice of Graph Convolutional Networks (GCN) \cite{kipf2016semi} and the use of its Multi-Relational Variant as a choice for this problem stems from the recent success of such models in learning over data that does not present itself in a regular grid like structure and yet can be modeled as a graph such as in social and biological networks.  Since a road scene can also be represented as a graph with nodes sharing multiple relations with other nodes, a GCN based model is apt for inferring overall node (object) behavior from a graph of interconnected relationships.

\textit{The decomposition of a dynamic on-road scene into its associated Interaction graph and the classification of the agent behavior by the MRGCN supervised over labels that are human-understandable (Lane Change, Overtake, etc.) form the main thesis of this effort.} Such behavior classification of agents in the scene finds immediate utility in downstream modules and applications. Recent research showcase results that understanding on-road vehicle behavior leads to better behavior planners for the ego vehicle \cite{meghjani2019context}. In \cite{meghjani2019context}, belief states over driver intents of the other vehicles where the intents take the labels ``Left Lane Change", ``Right Lane Change", ``Lane Keep" are used for a high-level POMDP based behavior planner for the ego vehicle. For example, the understanding that a car on the ego-vehicle's right lane is executing a lane change behavior into the current lane of the ego-vehicle can activate its ``Change to Right Lane" behavior operation option for path planning. Similarly, modeling an agent's behavior such as a car in a parked state can make the ego vehicle use feature descriptors of the parked car to update its state accordingly, which would not be possible if the active object was engaged in any other behavior. Understanding on-road vehicle behaviors also lend itself to very pertinent applications such as detecting and classifying traffic scene violations such as ``Overtaking Prohibited", ``Lane Change Prohibited." 

Our contributions are as follows:\\
% \begin{enumerate}
    % \item 
1) We propose a novel yet simple scheme for spatial behavior encoding from a sequence of single-camera observations using straight forward projective geometry techniques. This method of encoding spatial behaviors for agents is better than previous efforts that have used end-to-end learning of spatial behaviors \cite{mylavarapu2020accurate}.

\noindent2) We demonstrate the aptness of the proposed pipeline that directly encodes Spatio-temporal behaviors as an intermediate representation into the scene graph $G$, followed by the MRGCN based behavioral classifier. We do this by comparing with two previous methods \cite{mylavarapu2020accurate, jain2016structural} that are devoid of such intermediate Spatio-temporal representations but activate the behavior classifier on per frame spatial representations sequenced temporally.  Specifically we tabulate significant performance gain of at-least 25$\%$ on an average vis a vis \cite{jain2016structural} and 10$\%$ over \cite{mylavarapu2020accurate} on a variety of datasets collected in various parts of the world \cite{Geiger2013IJRR, huang2018apolloscape, 360LiDARTracking_ICRA_2019} and our own native dataset (refer Table \ref{table:main}).

\noindent3) We signify through a label deficient setup, the need for a neural-attention component that integrates with the MRGCN and further boosts its performance to nearly perfect predictions, as seen in Tables \ref{table:ablation} and \ref{table:complete}. Critically incorporating the attention function leads to high performance even in a limited training set, which the MRGCN without attention function cannot replicate. 

\noindent4) We also show seamless transfer of learning without further need to fine-tune across various combinations of datasets for the train and test split. Here again, we show better transfer capability of the current model vis a vis prior work, as shown in Table \ref{table:transfer_complete}.
% \end{enumerate}
% Through qualitative and quantitative performance gain achieved vis a vis with previous approaches and improved transfer learning performance across datasets, the efficacy of the proposed framework is vindicated.
\begin{figure*}
    \centering
    \includegraphics[trim=0 0 0 0,clip,  width=0.94\linewidth]{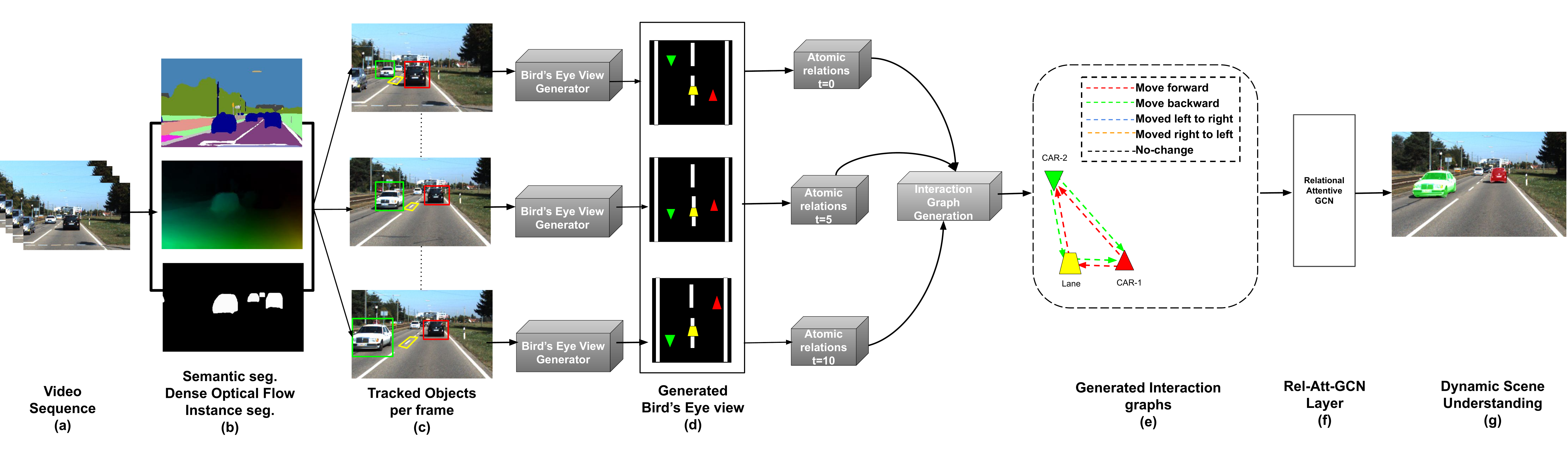}
    \caption{\scriptsize \textbf{Overall pipeline of our framework:} The input (a) to the pipeline is monocular image frames. Various object tracking pipelines are used to detect and track objects, as shown in (b). (c) denotes the tracked objects. Tracklets for each object are projected to 3D space, Bird's eye-view at each time step, as shown in (d). Spatial relations from Bird's eye view are used to generate Interaction graphs (e). This graph is passed through a Rel-Att-RGCN (f) to classify objects in the scene as shown in (g).}
    \label{fig:pipeline_full}
    % \vspace{-1mm}
\end{figure*}
% \section{Related Work}
% \subsubsection{Vehicle Behavior and Scene Understanding}
% The problem of vehicle scene understanding has traditionally has  been modeled eihter through a rule based approach \cite{geng2017scenario} or leverage a probalisitic model \cite{DBLP:journals/corr/abs-1803-07549,sivaraman2013looking,schulz2018interaction,jain2015car, mitrovic2005reliable,sabzevari2016multi,narayanan2019dynamic} for classification of driver behavior. \cite{schulz2018interaction,sabzevari2016multi} are concerned with predicting the future behavior  rather than classification. 
\section{Related Work}
\subsubsection{Vehicle Behavior and Scene Understanding}
The problem of on-road vehicle scene understanding is an important problem within autonomous driving. Most earlier works relied on multiple sensor-based data to solve this task. Rule-based \cite{geng2017scenario} and probabilistic modeling \cite{DBLP:journals/corr/abs-1803-07549,sivaraman2013looking,schulz2018interaction,jain2015car, mitrovic2005reliable,sabzevari2016multi,narayanan2019dynamic} where the goto approaches for classification of driver behavior with sensor data. Also, many of these works \cite{schulz2018interaction,sabzevari2016multi} were concerned only with predicting future trajectories rather than classification.  Herein, we chose a simpler and challenging set up to understand the scene and predict vehicle behaviors with observations from a single-camera in this work. While there are few works based on a single-camera data feed, they only focus on ego-centered predictions \cite{wu2019learning}. Here we focus on classifying other vehicles' behavior from an ego-vehicle perspective. Learning other vehicle behaviors can be helpful in behavior planning, state estimation, and applications relating to the detection of traffic violations over videos

\subsubsection{Graph based reasoning}
Graphs are a popular choice of data structures to model numerous irregular domains. With the recent advent of Graph Convolutional Networks (GCNs) \cite{kipf2016semi} that can obtain relevant node-level features for graphs, there is a widespread adaption of graph-based modeling of numerous computer vision problems such as in situation-recognition tasks \cite{lisituation}.  \cite{wang2019deep} encodes object-centric relations in an image using a GCNS to learn object-centric policies for autonomous driving. \cite{jain2016structural} and \cite{mylavarapu2020accurate} models objects in a video as Spatio-temporal graphs to make predictions of Spatio-temporal nature. It is common to model the temporal context of objects with recurrent neural nets and spatial context with a graph-based neural net. 

In this work, we focus on the task of on-road vehicle behavior and show that a proposed intermediate representation, called an Interaction graph, can yield better performance over working with a raw set of spatial graphs as done traditionally. This also portrays current models' incapacity to learn end-end and derive such useful features as with the Interaction graph from the raw spatial graphs. The closest works to ours are \cite{wu2019learning} and \cite{li2019learning}. They generate an affinity graph that captures actor-objects relationships. A simple GCN is then used to reason over this graph to classify ego-car action and not other vehicles. In our work, we use a richer multi-relational graph and a corresponding multi-relational GCN to work on the same. Further, we propose an attention based model that can leverage different relation types depending on the scene context.
\begin{figure*}
    \centering
    \includegraphics[trim=0 0 0 0,clip,  width=0.94\linewidth]{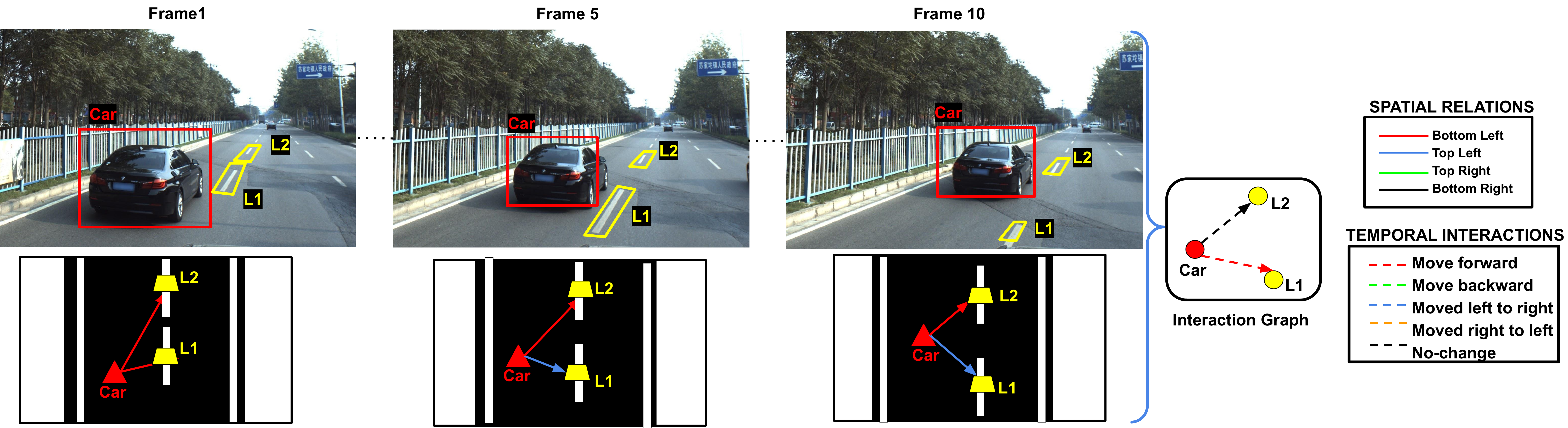}
    \caption{\scriptsize \textbf{
Temporal Interaction Graph Generation:} {The top row contains image frames at t=0, t=5, t=10 time step with tracked objects. The bottom row shows the corresponding bird's eye view. In the scene, a car and two lane-markings are tracked at each time step. The thick red, blue edges between car, L1 in first, and the final frame denote their spatial relations. The car moves forward with respect to L1 and has no relational change with L2, as shown with thick edges in the Interaction graph. Temporal relations are represented with dotted lines, while spatial relations are represented with thick edges. Corresponding color coding is shown in legends.}}
    \label{fig:graph_pipeline}
% \vspace{-2mm}
\end{figure*}

\section{Proposed Methodology}
Dynamic scene understanding requires well modeling of the different Spatio-temporal relations that may exist between various active objects in a scene. Towards this goal, we propose a pipeline that first computes a time-based ordered set of spatial relations for each object in the video scene. Secondly, it generates a multi-relational interaction graph representing the temporal evolution of the spatial relations between entities obtained from the previous step. Finally, it leverages a graph-based behavior learning model to predict behaviors of vehicles in the scene. Fig \ref{fig:pipeline_full} provides an overview of our proposed framework. 

Our proposed pipeline leverages and improves the data modeling pipeline introduced in \cite{mylavarapu2020accurate} (MRGCN-LSTM). Our pipeline's performance gains primarily stem from two of our contributions: (i) the \textit{interaction graph} that provides useful and explicit temporal evolution information of spatial relations. (ii) the proposed Multi-Relational Graph Convolutional Network (MRGCN) with our novel  \textit{multi-head relation attention function}, which we name Relation-Attentive-GCN (Rel-Att-GCN). 

\subsection{Spatial Graph Generation}
For dynamic scene understanding, we need to identify different objects in the scene and determine the atomic spatial relationships between them at each time-step. This phase of the pipeline closely follows the spatial scene graph generation step of \cite{mylavarapu2020accurate}. refer to the same for mode details.

\subsubsection{Object Detection and Tracking}
\label{sec:tracking}
Different \textit{vehicles} in the video frames are detected and tracked through \textit{instance segmentation} \cite{MaskRCNN} and \textit{per-pixel optical flow} \cite{pathakCVPR17learning} respectively. The instance-level segmentation of an object obtained from MaskRCNN are projected to the next frame and are associated with the highest overlapping instance using optical flows. Apart from tracking vehicles in the scene, static objects such as \textit{lane markings} are also tracked with \textit{semantic segmentation} \cite{rota2018place} to better understand changing relations among static and non-static objects. \cite{mylavarapu2020accurate} has shown that performance improvement can be obtained by leveraging a higher number of static objects in the scene.  

\subsubsection{Bird's Eye View} \label{sec:Birds_Eye_View}
The tracklets of objects obtained in the image space are re-oriented in the Bird's eye view (Top View) by projecting the image coordinates into 3D coordinates as described in \cite{DBLP:conf/cvpr/SongC14}. This reorientation facilitates determining spatial relations between different entities. Each object is assigned a reference point to account for the difference in heights. The reference is at the center for lane markings, and for vehicles, it is the point adjacent to the road. 
% For a given homogeneous co-ordinate $b = (x,y,1)$ in image space, we define $B = (B_x,B_y,B_z)$ in 3D co-ordinates as described in Eqn: \ref{eqn:birds}.
% \begin{align}
% \label{eqn:birds}
% \small
%     B = (B_x, B_y, B_z) = \frac{-hK^{-1}b}{\eta^{T}K^{-1}b}
% \end{align}
% IN Eqn: \ref{eqn:birds}, $K$ is the Camera intrinsic matrix, $h$ being the height of camera from ground plane or road plane. $\eta$ is defined as the normal to the ground plane and $b$, $B$ are co-ordinates of objects in image and 3D space respectively.

\subsubsection{Spatial relations}
At all $T$ frames of the video, the spatial relations between different entities are determined using their 3D positional information in the Bird's eye view. Specifically, the spatial relations are the four quadrants, \{\textit{top left, top right, bottom left, and bottom right}\}. For a subject entity, $i$, its spatial relation with an object entity, $j$ at time-step, $t$ is denoted as $S^t_{i,j}$. 

% $S^t_{i,j}$ corresponds to the atomic edge between node $i$ and node $j$ at frame number $t$. We partition the regions around the subject node, $i$ into four quadrants viz: \{\textit{top left, top right, bottom left and bottom right}\}. The value for $S^t_{i,j}$ is determined by the quadrant in which object, $j$ lies w.r.t to subject, $i$. Thus $S^t_{i,j} \in \mathcal{R}_s$, with $\mathcal{R}_s \in \{1,..,4\}$ as the set of atomic relations of interest corresponding to the four quadrants. These spatial relations are best illustrated in Fig. \ref{fig:graph_pipeline} in legend \textit{Atomic Relations} . 

%The scene graph \label{sec:SceneGraph} the scene graph is generation of a \textit{Scene Graph}, we observe objects of interest (\textit{Vehicles and Lane-Markings}) through $T$ time steps. Every tracked object in the video is considered a node in the graph. For each node, we obtain it's 3D positions from \ref{sec:Birds_Eye_View} through $T$ time steps.
%We quantize these evolving spatial relations between every two nodes in the graph at each time $t$. Formally, $r_{i,j}^t$ depicts the saptial relation between node $i$ and node $j$ at time $t$. The quadrant in which node $i$ lies with respect to node $j$ at time $t$ determines the value of $r_{i,j}^t$.  Here, $r_{i,j}^t \in R_s$, where $R_s$ = \{\textit{top-left, bottom-left, top-right, bottom-right}\}. These spatial relations are observed through $T$ time steps. 

\subsection{Temporal Interaction Graph Generation}
\label{sec:SceneGraph}
Modeling object interactions as a time-based ordered set of spatial graphs is a popular approach in many Spatio-temporal problems.
such as  \cite{geng2020spatio,ji2020action,herzig2019spatio,wang2018videos} where spatial graphs are constructed over object interactions   to classify actions in a video stream. \cite{jain2016structural, mylavarapu2020accurate} use a similar approach for predicting on road object behaviors. We found that it is harder for models learned with such data to learn some of the simple temporal-evolution behaviors needed for the end task, specifically in our problem of interest. In our problem of focus, behavior prediction, it is important for the model to learn the nature of some simple temporal evolution of interaction between entities such as \textit{move-forward, move-backward, moved left to the right, moved right to the left, no-change}. However, we found that explicitly modeling such information was highly beneficial. A simple rule-based model with such interaction information outperformed learned models on the primitive information at the level of spatial relations. Having motivated by this insight, we propose a way to define an Interaction graph with temporal evolution information and a new model that can benefit from such information.

% \textcolor{blue}{The interaction graph is generated based on a set of rules defined over the edges between objects}. 
The Interaction graph summarizes the temporal evolution of spatial relations from $T$ frames into a \textit{single multi-relational graph with temporal relations}, $R_{d}$ = \{\textit{move-forward, move-backward, moved left-to-right, moved right-to-left, no-change}\}. The edge, $E_{i,j} \in R_{d}$, denotes the temporal relation between subject entity, $i$ and object entity, $j$. $E_{i,j}$ is computed by deterministic rule based pipeline, that takes in their spatial relations over $T$ frames, i.e, \{$S^1_{i,j}, S^2_{i,j},...,S^T_{i,j}$\}.

For example, an object entity, $j$ that is initially in the \textit{bottom-left} quadrant with respect to subject entity $i$, changes it's atomic spatial relation to \textit{top-left} with respect to $i$ at some time $t$, will have its temporal relation $E_{i,j}$ as \textit{move-forward}. Similarly, an object entity, $j$ that is initially in \textit{bottom-left} quadrant changes to \textit{bottom-right} quadrant with respect to subject $i$ at some time $t$, will have temporal relation $E_{i,j}$ as \textit{moved left-to-right}. Since these temporal relation annotated edges have a proper direction semantics, we can't treat the graph is undirected. Thus, we also introduce inverse edges for complementary relations suchs as \textit{moved forward and backward} and \textit{moved left-to-right and moved right-to-left}. An overview of the temporal interaction graph generation is presented in Fig. \ref{fig:graph_pipeline}. 

\subsection{Behavior Prediction Model}
We propose a Multi-relational Graph Convolution Network (MRGCN) with a relation-attention module that conditioned on the scene, automatically learns relevant information from different temporal relations necessary to predict vehicle behaviors. 

\subsubsection{Multi-Relational Graph Convolution Networks}
\label{sec:MRGCN}

Recently Graph convolution Networks \cite{kipf2016semi}
has become the popular choice to model graph-structured data. We model our task of maneuver prediction by using a variant of Graph Convolutional Networks, Multi-Relational Graph Convolutional Networks (MRGCN) \cite{schlichtkrull2018modeling} originally proposed for knowledge graphs with multiple relation types. MRGCN is composed of multiple graph convolutional layers, one for each relation between nodes. A Graph Convolution operation for a relation $r$ here is a simple neighborhood-based information aggregation function. In the MRGCN, information obtained from convolving over different relations is combined by summation.

Let us formally define the temporal Interaction Graph as $G = (V,E)$ with vertex set $V$ and edge set, $E$, where $E_{i,j} \in R_{d}$ is an edge between node $i$ and $j$. The $i^{th}$ node feature obtained from a graph convolution over relation, $r$ in $l^{th}$ layer is defined as follows:
\begin{align}
\label{eqn:MR-GCN1}
    h^l_{r}[i] &= \sum_{j \in \mathcal{N}_r[i]} \dfrac{1}{c_{r}[i]} W_r^l h^{l-1}[j]
\end{align}
where, $\mathcal{N}_r[i]$ denotes set of neighbour nodes for $v_i$ under relation $r$, $\mathcal{N}_r[i] = \{ j \in V \ | \ {E_{j,i}} = r\}$ and $c_{r}[i] = | \mathcal{N}_r[i] |$ is a normalization factor. Here, $W_r^l \in \mathcal{R}^{d'*d}$ is the weights associated with relation $r$ in the $l^{th}$ layer of MR-GCN;  $d',d$ are dimensions of $(l-1)^{th}$ and $l^{th}$ layers of MRGCN. 

Neighborhood information aggregated from all the relations are then combined by a simple summation to obtain the node representation as follows:
\begin{align}
\label{eqn:MR-GCN2}
    h^l[i] &= ReLU(W_s^l h^{l-1}[i] + \sum_{r \in R_d} h^{l}_r[i])
\end{align}
where, the first terms correspond to the node information (self-loop) and $W_s \in \mathcal{R}^{d'*d}$ is the weight associated with self-loop. To account for the nature of the entity (active or passive), we learn entity embeddings, $\mathcal{E}_i \in \mathbb{R}^{|O|*d}$ for node $v_i$, where $O = \{Vehicles,\ Lane \ Markings\}$. The input to the first layer of the MRGCN, $h^0[i]$, is the embedding $\mathcal{E}_i$ based on type of node $i$.    

\subsubsection{Relation-Attention MRGCN (Rel-Att-GCN)}
\label{sec:attention}
The MRGCN defined in Eqn: \ref{eqn:MR-GCN2} treats information from all the relations equally, which might be a sub-optimal choice to learn discriminative features for certain classes. Motivated by this, we propose a simple attention mechanism that scores the node information's importance along with individual neighborhood information from each relation.

The attention scores, $\alpha$ are computed by concatenating the information from the node ($h^{l-1}$) and its relational neighbors ($h_r^l$)  and transforming it with a linear layer to predict scores for each component. The predicted scores are softmax normalized. The attention scores are computed as defined below.
\begin{align}
\label{eqn:MR-GCN3}
\small
    \alpha^l[i] = softmax([ h^{l-1}[i]\parallel h^l_1[i]\parallel h^l_2[i]...\parallel h^l_{|R_d|} ]W_u^l)
\end{align}
where, $\parallel$ represents concatenation and $\alpha^l[i] \in \mathcal{R}^{|R_d+1|}$ with $W_u^l$ being the linear attention layer weights. These probabilities  depict the importance of a specific relation conditioned on that node and its neighborhood. 

The attention scores are used to scale the node and neighbor information accordingly to obtain the node representation as follows.
\begin{align}
\label{eqn:attention}
    h^l[i] = ReLU(\alpha_{node}^l[i] h^{l-1}[i] + \sum_{r \in R_d}\alpha_r^l[i]  h^{l}_r[i])
\end{align}
where, $\alpha^{l}_{node}$ is the attention score for self-loop.
The node representation obtained at the last layer is used to predict labels, and the model is trained by minimizing the cross-entropy loss.

\section{Experiment and analysis}
\subsection{Dataset}
Numerous datasets have been released in the interest of solving problems related to autonomous driving. We choose four datasets for evaluating our framework, of which three are publicly available: ApolloScapes \cite{huang2018apolloscape}, KITTI \cite{Geiger2012CVPR}, Honda Driving dataset \cite{360LiDARTracking_ICRA_2019} and one is a proprietary Indian dataset. These datasets provide hours of driving data with monocular image feed in various driving conditions. We use the same dataset Train/Test/Val splits from \cite{mylavarapu2020accurate} for Apollo Scape, KITTI, and Indian dataset and extend the setup to Honda dataset and manually annotated accordingly for our task. 

\noindent1) Apollo Dataset: 
We choose Apollo-scapes as our primary dataset as it contains a large number of driving scenarios that are of interest. It includes vehicles depicting overtake and lane-change behaviors. The dataset consists of image feed collected from urban areas and contains various objects such as Cars, Buses, etc. The final dataset used here contains a total of 4K frames with multiple behaviors.
\noindent2) KITTI Dataset:
It consists of images collected majorly from highways and wide open-roads, unlike Apollo-scapes. We select 700 frames from \textit{Tracking Sequences} 4, 5, and 10 for our purpose. These chosen sequences contained a variety of driving behaviors compared to the rest.
\noindent3) Honda Driving Dataset:
This dataset consists of multiple datasets in itself. From among them, we choose the H3D dataset for our task as it contained lane change behaviors. H3D comprises driving in urban city conditions where lane changes are prominent. We excluded overtaking behavior here due to its fewer occurrences in the dataset. It consists of a total of 1.5K frames.
\noindent4) Indian Dataset
Although the datasets mentioned above are widely used and include wide vehicle behaviors, they mostly contain standard vehicle frames only. To showcase models' transfer learning capabilities on less standard vehicles, we also use an Indian driving dataset that includes vehicles such as auto-rickshaw, trucks, tankers, etc. This dataset contains 600 frames.

Class labels: The vehicle behaviors predicted in these datasets are: (i) Moving Away from Us (MAU), (ii) Moving Towards Us (MTU), (iii) Parked (PRK), (iv) Lane Change from Left-right (LCL), (v) Lane Change from left-Right (LCR) and (vi) Overtake (OVT).

\subsection{Experimentation details}
All the models, both learning and rule-based, use the same pre-processing steps to identify and track objects in the scene, as explained in section: \ref{sec:tracking} for $T =$ 10 time-steps (frames). The Spatial graphs at each time frame are constructed by considering a maximum of 10 vehicles in the scene nearest to the ego-vehicle for classifying them. Then, an \textit{Interaction graph} is independently generated from the set of $T$ temporally ordered spatial graphs. The MRGCN used is identical in both models MRGCN and Rel-Att-GCN. Note that the input and output dimensions of attention are equal. We empirically found that using 3 layers of MRGCN with dimensions 64, 32, and 6 (number of classes) respectively works best for our task. In the case of \textit{Rel-Att-GCN}, simple attention is applied over the output of MRGCN for each node individually with 2 heads. Outputs of the heads are concatenated across relations and projected back to the MRGCN layer's output dimension with a linear transformation. We found adding skip connection from every layer $l$ to $(l+2)^{th}$ layer to be beneficial. 

The inference time for the models: MRGCN-LSTM, MRGCN, and Rel-Att-GCN averaged over 1K graphs are 0.02, 0.03, and 0.04 seconds respectively. Note that the latter two models inference time also includes the creation of the Interaction graph. All the training and testing was done on a single Nvidia Geforce Gtx 1080 GPU. More details regarding the implementation can be found on our project website,  \footnote{\href{https://github.com/ma8sa/Undersrtanding-Dynamic-Scenes-using-MR-GCN.git}{code:https://github.com/ma8sa/Undersrtanding-Dynamic-Scenes-using-MR-GCN.git}}.

\label{sec:results}
\begin{table}[t]
\vspace{2mm}
\label{table:main}

\renewcommand{\arraystretch}{1.3}
    \centering
    \begin{adjustbox}{width=0.48\textwidth,center}
    \begin{tabular}{|c|c||c|c|c|c|c|c|}
    \hline
    %  \textbf{\shortstack{Method \\ Used}} & \shortstack{Moving \\ away \\ from us } & \shortstack{Moving \\ towards \\ us} & Parked & \shortstack{Lane \\ change \\ L-R} & \shortstack{Lane \\ change \\ R-L} & \shortstack{Over \\ take}  \\
    \textbf{\shortstack{Graph used}} & \textbf{\shortstack{Methods}} & \shortstack{Moving \\ away \\ (MAU)} & \shortstack{Moving \\ towards \\ (MTU)} & Parked & \shortstack{Lane change \\ L-R (LCL)} & \shortstack{Lane change \\ R-L (LCR)} & \shortstack{Overtake}  \\
    \hline
    Temporally & St-RNN \cite{jain2016structural}  & 76 & 51 & 83 & 52 & 57 & 63  \\ 
    \shortstack{ordered set \\ of Spatial graphs}  & MRGCN-LSTM \cite{mylavarapu2020accurate}& 85 & 89 & 94 & 84 & 86 & 72  \\ 
    \hline 
    Spatio-Temporal & Rule Based & 90 & \textbf{99} & \textbf{98} & 81 & 87 & \textbf{90}  \\ 
    % Spatio-temporal & \textbf{MRGCN}    & 94 & 95 & 94 & 97 & 93 & 86  \\
    % Interaction graph & \textbf{\shortstack{Rel-Att-GCN}} & \textbf{95} & \textbf{99} & \textbf{98} & \textbf{97} & \textbf{96} & \textbf{89}  \\
     Interaction & MRGCN \cite{schlichtkrull2018modeling, mylavarapu2020accurate}& 94 & 95 & 94 & 97 & 93 & 86  \\
     graphs & \textbf{\shortstack{Rel-Att-GCN}} & \textbf{95} & \textbf{99} & \textbf{98} & \textbf{97} & \textbf{97} & 89  \\
    \hline
    \end{tabular}
    \end{adjustbox}
% \caption{\scriptsize Comparison of our framework between various models along with baselines tested on Apollo Scapes dataset.Values provided are accuracies in percentage.}
    \caption{\scriptsize {Vehicle behavior prediction on Apollo Scape dataset.}}
    
    \vspace{-4mm}
\end{table}
\subsection{Baseline Comparisons}
In Table: \ref{table:main}, we compare our models with Spatio-temporal approaches as well as a rule-based method on the Interaction graph. Table \ref{table:main} reports class-wise Recall scores of these methods. The results reported here in all the tables are averaged over 5 runs.

\subsubsection{Spatio-Temporal approaches}
To depict the importance of encoding Spatio-temporal information as an Interaction graph, we compare our model with Structural-RNN \cite{jain2016structural} and MRCGN-LSTM \cite{mylavarapu2020accurate} that processes a time based ordered set of spatial graphs. Structural-RNN (St-RNN) encodes the spatial representation for each frame in a graph and then reasons over the temporal evolution of these graphs by feeding it to a Recurrent Neural Network. We adapt St-RNN's pipeline to our problem by replacing humans and objects in their model with vehicles and stationary landmarks, respectively. A similar methodology is employed for MRGCN-LSTM \cite{mylavarapu2020accurate}.

We show a quantitative comparison with our pipeline/model variations in Table \ref{table:main}. St-RNN that doesn't have any GCN components fare the worst. In Table \ref{table:main}, we observe that our method outperforms the traditional temporal based approach, St-RNN, by a significant margin. The gap is even more prominent when comparing the harder classes such as lane changing and overtaking, where we observe an average difference of 40\% and 26\%, respectively. A comparison between MRGCN-LSTM that uses a set of spatial graphs vs. MRGCN that uses the proposed interaction graph clearly shows the benefit of the proposed interaction graph. MRGCN outperforms its counter that learns in an end-end manner. This shows how such simple inherent behaviors are still hard for GCNs to learn. Further, with the addition of the attention mechanism, the Rel-Att-GCN model achieves an additional absolute 3-4\% improvement on a few hard classes.

%includes Human activity detection, Human motion modeling and forecasting, \textcolor{red}{Driver maneuver anticipation. As Driver anticipation classifies only ego-vehicle behaviour, we consider Human activity detection for our comparison due to it's similarity with our method in terms of modelling objects and relations between them.} Similar to our non-static (Vehicles) and static objects (Lane Markings), St-RNN uses Humans and static-objects to model relations for activity recognition. Hence, we use Vehicles as Humans and Lane Markings as Objects in their architecture. Based on spatial information obtained at each time step $t$, we pose our relational information between Vehicle-Vehicle, Vehicle-Lanes and Lanes-Lanes as their Human-Human, Human-Object and Object-Object interactions and train it to predict vehicle behaviours. Based on spatial relation $r_{i,j}$ between objects $i$ and $j$, and the nature of objects (static or non-static), we define an embedding at each time step $t$ which is given as input to the St-RNN. 

\subsubsection{Rule Based Baseline}
\begin{table}[!t]
\vspace{2mm}
\renewcommand{\arraystretch}{1.3}
    \centering
    \begin{adjustbox}{max width=\linewidth}
    \begin{tabular}{|c||c|c|c||c|c|c|}
    \hline
    \textbf{Model} & \multicolumn{3}{|c||}{\textbf{Rel-Att-GCN}} & \multicolumn{3}{|c|}{\textbf{Rule-Based}} \\
    \hline
     \shortstack{perf. measure}&  precision & recall & F1 & precision & recall & F1 \\
    \hline
    % MAU & 97 & 95 & 96   & 97 & 91 & 94 \\
    MAU & 97 & 95 & \textbf{96}   & 97 & 91 & 94 \\
    % MTU & 94 & 100 & 97      & 100 & 99 & 99 \\
    MTU & 95 & 99 & 97      & 100 & 99 & \textbf{99} \\
    % PRK & 100 & 98 & 99      & 100 & 99 & 99 \\
    PRK & 100 & 98 & \textbf{99}      & 100 & 99 & \textbf{99} \\
    % LCL & 94 & 98 & 96   & 96 & 81 & 88 \\
    LCL & 94 & 97 & \textbf{95}   & 96 & 81 & 88 \\
    % LCR & 97 & 98 & 97   & 97 & 88 & 92 \\
    LCR & 97 & 97 & \textbf{97}   & 97 & 88 & 92 \\
    % OVT & 72 & 88 & 79   & 36 & 90 & 52 \\
    OVT & 71 & 89 & \textbf{79}   & 36 & 90 & 52 \\
    \hline
    % micro avg &  97 &	97 & 97	 & 	 95	& 95	& 95 \\
    micro avg &  97 &	97 & 97	 & 	 95	& 95	& 95 \\
    % macro avg  &  92 & 96 &	94 &  88	& 91	& 87 \\
    macro avg  &  92 & 96 &	94 &  88	& 91	& 87 \\
    % weighted avg & 0.97 & 0.97 & 0.97 &  0.97 & 0.95 & 0.95 \\
    \hline
    \end{tabular}
     \end{adjustbox}
    \caption{\scriptsize Rel-Att-GCN Vs Rule-Based model on ApolloScape dataset. \newline Both the models use the proposed Interaction graph.}
    \vspace{-5mm}
    \label{table:precision}
\end{table}

To showcase the effectiveness of an information-rich Interaction graph over traditional Spatio-temporal modeling with a set of spatial graphs, we propose a rule-based approach to infer over the Interaction graph as one of our baselines. We use the Interaction Graphs generated by our pipeline (ref section:\ref{sec:SceneGraph}) and employ an expert set of rules carefully framed to classify between behaviors. The deterministic classification is a simple max function over different relations a vehicle is associated with.

Relational behavior with majority count decides to which class the object belongs to from the following, \{\textit{moving away, moving towards us, and lane changes}\}. For example, a node having the highest count for behavioral relation \textit{moved left to right} would have its class as Lane change (Left to right). To obtain classification for \textit{overtake} behavior, we iterate over all pair of vehicles, $i$ and $j$, that are not classified as \textit{parked} or \textit{moving towards us} in the first iteration and then we observe if there exists $e_{i,j}=$ \textit{move-forward}, in which

The quantitative comparison in Table \ref{table:main}, clearly depicts the advantage of a rich temporal Interaction graph over models that directly utilize a set of spatial graphs, especially in the complex behavior class overtaking where it shows an 18 \% and 27 \% over MRGCN and St-RNN respectively. Though the rule-based model performs well compared to the Spatio-temporal approaches, it falls short against the learning-based models trained on the Interaction graph. The rule-based model is not powerful enough, especially on lane-change classes. This clearly explains the need for a learnable model to learn complicated patterns. On a closer look in Table \ref{table:precision}, it is clear that the rule-based method is not consistently better, especially on precision and recall metrics of Overtake and lane-change classes, respectively. The proposed Rel-Att-GCN clear outperforms the rule-based model on aggregate Micro and Macro average scores.
\begin{table}[h]
\renewcommand{\arraystretch}{1.3}
    \centering
    % \vspace{2mm}
    \begin{adjustbox}{max width=\linewidth}
    \begin{tabular}{|c|c|c||c|c||c|c|}
    \hline
    \textbf{Train Ratio} & \multicolumn{2}{|c||}{0.05} & \multicolumn{2}{|c||}{0.1} & \multicolumn{2}{|c|}{0.2}  \\
    \hline
    \textbf{Method} & MRGCN & \shortstack{Rel-Att\\GCN} & MRGCN & \shortstack{Rel-Att\\GCN} & MRGCN & \shortstack{Rel-Att\\GCN} \\
    \hline
    MAU & 61 & \textbf{94} & 91 & \textbf{95} & 89 & \textbf{97} \\
    MTU & 47 & \textbf{99} & 75 & \textbf{99} & 85 & \textbf{99} \\
    PRK & 90 & \textbf{98} & 86 & \textbf{98} & 86 & \textbf{98} \\
    LCL & 36 & \textbf{98} & 69 & \textbf{98} & 89 & \textbf{98} \\
    LCR & 54 & \textbf{94} & 73 & \textbf{95} & 88 & \textbf{96} \\
    OVT & 50 & \textbf{60} & 58 & \textbf{65} & 76 & \textbf{77} \\
    \hline
    \end{tabular}
     \end{adjustbox}
     \caption{\scriptsize {Recall on ApolloScape dataset for different amount of training data}}
    % \caption{\scriptsize We provide results on different models trained on different ratios of the train data. Values provided are accuracies.}
    \vspace{-5mm}
    \label{table:ablation}
\end{table}
\subsection{Analysis of Relation-Attention MRGCN (Rel-Att-GCN)}
\begin{figure}[!ht]
    \centering
    \vspace{-4mm}
    \includegraphics[trim=0 0 0 0,clip,  width=0.94\linewidth]{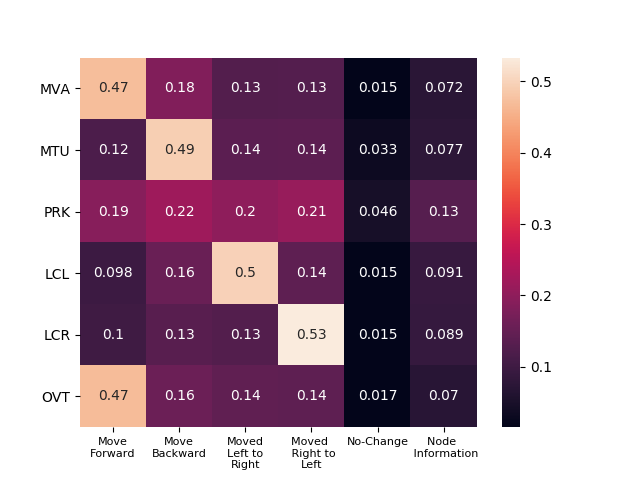}
    \caption{\scriptsize Figure shows attention scores between class labels (rows) and types of spatio-temporal interactions (columns). Higher scores for a particular relation indicates a higher dependence of the class on that particular relation.
    % Attention scores obtained from attention heads are normalized across relations and denoted in the figure. 
    % \textcolor{red}{The rows correspond to the classes while columns denote the relations. Higher scores for a particular relation shows higher dependence of the class on that particular relation.}
    % Abbreviations for class labels are, MVA : moving away from us, MTU: moving towards us, PRK: parked, LCL: lane change left to right, LCR: lane change right to left, OVT: overtaking 
    }
    \vspace{-2mm}
    \label{fig:attention_map}
\end{figure}

The Rel-Att-GCN model is the MRGCN model with an additional attention component. The proposed Attention function factors into account that different types of relations in the interaction graph may have different relevancy to predict different classes.  The varying importance of the relations in classifying vehicle behaviors can be visualized by analyzing normalized attention scores across relations for each class. One such visualization for the Apollo Scape dataset is depicted in Fig. \ref{fig:attention_map}. Higher values in each class (row) denote higher importance given by the attention function to that particular relation (column) to predict that class. The attention map clearly shows how classes such as overtake and lane changes depend on \textit{moving forward} and \textit{moved left to right} or \textit{right to left} respectively. Despite how both MVA and OVT classes have high probability mass on move-forward relation, they can distinguish themselves based on the attention score spread over other relations.

Such reasoning helps the network to learn effectively under label scarcity, as given in Table \ref{table:ablation}. Herein, we report results for models trained with 5\% , 10\% and 20\% of training data. Rel-Att-GCN is able to show fidelity even when only 5\% percent of data is present in contrast to a normal MR-GCN trained on the same interaction graph, which finds it difficult to learn from the smaller dataset. A similar trend is observed as we increase the size of training data to 10\% and 20\% of the actual dataset. In Table \ref{table:complete}, where the model is learned with 70\% data, we observe that Rel-Att-GCN achieves superior performance across all datasets when compared with plain MR-GCN as well as temporal based methods.    

%\textcolor{blue}{Attention denotes importance of different relations for every object in the scene. This helps the network to give higher importance or weight to relations that determine the label for each node}.an N x d output from all relations are concatenated to make N x r x d. d x 1 attention Weight is applied to project to N x r x 1. Softmax on this gives weight to all relations. 

\subsection{Transfer Learning}
\begin{table*}[!ht]
\renewcommand{\arraystretch}{1.3}

    \centering
    % \begin{adjustbox}{width=\linewidth}
    \vspace{2mm}
    \begin{tabular}{|c||c|c|c||c|c|c||c|c|c|}
    \hline
    \textbf{Trained on} & \multicolumn{9}{|c|}{\textbf{Apollo}} \\
    \hline
    \textbf{Tested on} & \multicolumn{3}{|c||}{\textbf{Honda}} & \multicolumn{3}{|c||}{\textbf{KITTI}} & \multicolumn{3}{|c|}{\textbf{Indian}}  \\
    \hline
    Method  & \shortstack{MRGCN\\LSTM} & MRGCN & \shortstack{Rel-Att\\GCN} & \shortstack{MRGCN\\LSTM} & MRGCN & \shortstack{Rel-Att\\GCN} & \shortstack{MRGCN\\LSTM} & MRGCN & \shortstack{Rel-Att\\GCN}   \\
    % Tested on & Honda & & Honda \\
    \hline

    Moving away from us     & 55 & 92 & \textbf{92}   & 99 & 99 & \textbf{99}  & 99 & 98 & \textbf{99}  \\
    Moving towards us       & 79 & 60 & \textbf{92}   & 98 & 98 & \textbf{98}  & 93 & 94 & \textbf{97} \\
    Parked                  & 91 & 65 & \textbf{99}   & 99 & 98 & \textbf{99}  & 99 & 95 & \textbf{99} \\
    Lane-Change(L)          & 65 & 73 & \textbf{94}   & - & - & - & - & - & - \\
    Lane-Change(R)          & 87 & 83 & \textbf{92}   & - & - & - & - & - & - \\
    
    % Overtake            & \textcolor{red}{60} & \textcolor{red}{65} & \textcolor{red}{51} &\textcolor{red}{70} & \textcolor{red}{57} \\
    % Overtake            & \textcolor{red}{70} & \textcolor{red}{57} & \textcolor{red}{51} &\textcolor{red}{-} & \textcolor{red}{-} & \textcolor{red}{-} \\
    
    \hline
    \end{tabular}
    % \end{adjustbox}
    \caption{\scriptsize Transfer learning results: We train the models on Apollo Scapes dataset and test on Honda, KITTI and Indian datasets. 
    % Accuracies of each class are depicted from 4th row onwards . 
    % The second row denotes the datasets and third row denotes the method used. 
    % All three pre-trained models, MRGCN-LSTM, MRGCN and REL-ATT-GCN  are tested on all three datasets and all accuracies reported are in percentages. 
    }
    % \vspace{-6mm}
    \label{table:transfer_complete}
\end{table*}

\begin{table*}[!ht]
\renewcommand{\arraystretch}{1.3}
    \centering
    \begin{adjustbox}{max width=\linewidth}
    % \vspace{2mm}
    \begin{tabular}{|c||c|c|c||c|c|c||c|c|c||c|c|c|}
    \hline
    \textbf{\shortstack{Trained and\\Tested on}} & \multicolumn{3}{|c||}{\textbf{Apollo}} &  \multicolumn{3}{|c||}{\textbf{Honda}} & \multicolumn{3}{|c||}{\textbf{KITTI}} & \multicolumn{3}{|c||}{\textbf{Indian}}\\
    \hline
    \textbf{Method} & \shortstack{MRGCN\\LSTM} & MRGCN & \shortstack{Rel-Att\\GCN}  & \shortstack{MRGCN\\LSTM} & MRGCN & \shortstack{Rel-Att\\GCN}  & \shortstack{MRGCN\\LSTM} & MRGCN & \shortstack{Rel-Att\\GCN}  & \shortstack{MRGCN\\LSTM} & MRGCN & \shortstack{Rel-Att\\GCN} \\
    % Tested on & Honda & & Honda \\
    \hline

    Moving away from us & 85 & 94 & \textbf{95}   & 83 & 97 & \textbf{99}   & 85 & 92 & \textbf{98}   & 85 & 90 & \textbf{97}  \\
    Moving towards us   & 89 & 95 & \textbf{99}   & 79 & 86 & \textbf{90}   & 86 & 91 & \textbf{98}   & 74 & 91 & \textbf{97}  \\
    Parked              & 94 & 94 & \textbf{98}   & 85 & 88 & \textbf{99}   & 89 & 95 & \textbf{99}   & 84 & 93 & \textbf{99}  \\
    Lane-Change(L)      & 84 & 97 & \textbf{97}   & 75 & 91 & \textbf{91}   & - & - & -      & - & - & - \\
    Lane-Change(R)      & 86 & 93 & \textbf{97}   & 60 & 81 & \textbf{85}   & - & - & -      & - & - & - \\
    Overtake            & 72 & 86 & \textbf{89}   & - & - & -      & - & - & -      & -  & - & - \\
    
    % Overtake            & \textcolor{red}{60} & \textcolor{red}{65} & \textcolor{red}{51} &\textcolor{red}{70} & \textcolor{red}{57} \\
    % Overtake            & \textcolor{red}{70} & \textcolor{red}{57} & \textcolor{red}{51} &\textcolor{red}{-} & \textcolor{red}{-} & \textcolor{red}{-} \\
    
    \hline
    \end{tabular}
    \end{adjustbox}
    \caption{\scriptsize Performance of methods on different datasets. The models here are trained and tested on the same dataset. 
    % We report accuracies for different models trained and tested on multiple datasets. The top row denotes the dataset used for training and testing on the same with a validation split. 
    % The second row denotes the method used. 
    % 3 Columns under each dataset denote the class wise accuracies for all the three models , MRGCN-LSTM, MRGCN, REL-ATT-GCN on that dataset. All the accuracies are in percentage. 
    }
    \vspace{-6mm}
    \label{table:complete}
\end{table*}

To showcase our proposed pipeline's generality, we trained the model only on the Apollo dataset and tested it on validation sets of Honda, KITTI, and Indian datasets. At the testing phase, we removed the classes which were not present in corresponding datasets. As the proposed pipeline does not rely upon any visually learned features, we can achieve fidelity across all datasets, as seen in Table \ref{table:transfer_complete}. Evaluation results for the model trained and tested on validation sets of the same dataset can be found in Table \ref{table:complete}. From a comparison between Table \ref{table:transfer_complete} and Table \ref{table:complete}, the transfer learning model though not better than models that are trained and tested on the same dataset, is on par with them. Notably, in the Honda dataset, the Rel-Att-MRGCN performs better in transfer for all classes than the model trained and tested on Honda. We attribute this behavior to high variation present in the Apollo dataset, which the other datasets lack.
\subsection{Qualitative }
\begin{figure*}
\vspace{2mm}
% using graphs  Behaviour/scene classification by learning over graphs is gaining traction in the computer vision community. 
    \centering
    \includegraphics[trim=0 0 0 0,clip,  width=0.94\linewidth]{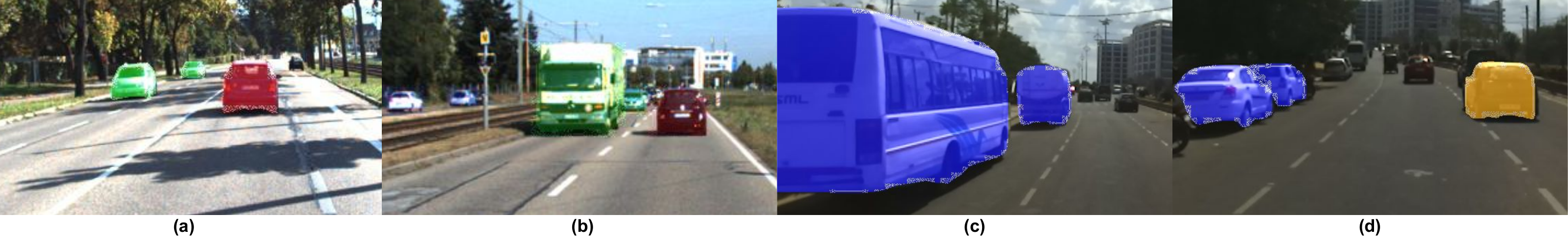}
    \caption{\scriptsize 
    % The figure depicts an accurate behavior classification of standard and non-standard vehicles. 
    Sub-figures (a) and (b) showcases prediction on standard vehicles from KITTI dataset. Whereas in (c), we can see Petrol Tanker and Bus's behavior predictions from the Indian dataset that shows object class-agnosticism. In (d), we can see a complex lane changes behavior prediction.}
    \label{fig:kitti_indian}
\end{figure*}

\begin{figure*}
% using graphs  Behaviour/scene classification by learning over graphs is gaining traction in the computer vision community. 
    \centering
    \includegraphics[trim=0 0 0 0,clip,  width=0.94\linewidth]{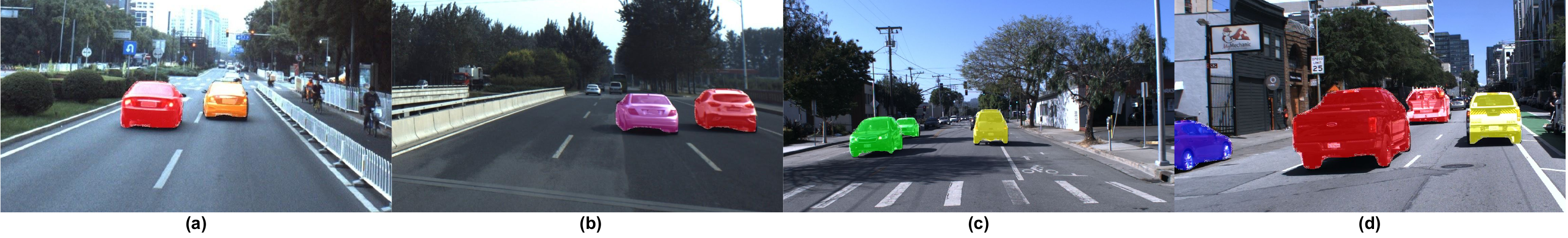}
    \caption{\scriptsize The figure shows multiple scenarios depicting various behaviors. (a), (b) are samples from ApolloScapes dataset while (c) and (d) are from Honda dataset. In (a), (c), and (d), we observe the model accurately classifying lane change, both left to right and right to left. (b) depicts a case where the magenta car overtakes the red car.}
    \vspace{-3mm}
    \label{fig:apollo_honda}
\end{figure*}

A video demonstration of the qualitative performance of our model on different datasets can be found here\footnote{\href{https://youtu.be/TT4J-uH4xqI}{Video : https://youtu.be/TT4J-uH4xqI}}. In Fig. \ref{fig:kitti_indian} and \ref{fig:apollo_honda} we showcase few qualitative results from different video snapshots. We follow a consistent convention for color-coding to depict behaviors. Red depicts vehicles \textit{Moving Away From Us}, Green for \textit{Moving towards Us} and Blue for \textit{Parked} vehicles and Yellow and Orange depict \textit{Lane Change Left to right} and \textit{Lane Change Right to left} respectively while Magenta corresponds to \textit{overtaking} vehicles.  

In Fig. \ref{fig:kitti_indian}, sub-figures (a) and (b) show instances of vehicles \textit{Moving Away From Us  and Moving towards} on the KITTI dataset. On the same Fig. \ref{fig:kitti_indian}, sub-figures (c) and (d), showcase results from the Indian Dataset, wherein image (c), we see a bus and truck parked and in (d) we see \textit{Lane change} behavior depicted by the car on the right. 

In Fig. \ref{fig:apollo_honda} (a) we see a car changing lane and in Fig. \ref{fig:apollo_honda} (b) we observe a car classified as overtaking. Fig. \ref{fig:apollo_honda} (c) and (d) show fidelity of our pipeline in traffic scenarios. In Fig.\ref{fig:apollo_honda} (c) we observe a car \textit{changing lane} and merging into the road on the right and two cars \textit{coming towards us}, in (d) we see a car changing a lane (on the right), a car parked on the left and two pickup trucks \textit{moving away} from us. The qualitative results validates the proposed model's near-perfect classification and generalizability across datasets even in the presence of less (or not) observed test vehicles. 

\section{CONCLUSIONS}
This paper proposed a novel pipeline for on-road vehicle behavior understanding and classification. It decomposed an evolving dynamic scene into a multi-relational Interaction graph whose nodes are the agents/actors in the scene, and edges are Spatio-temporal encodings that signify the agents' spatial behaviors. The interaction graph was further acted upon by a Multi-Relational Graph Convolution Network (MRGCN) to learn and classify the vehicle's overall behavior. The key takeaway is this two-stage classification that showed much-improved performance over end-end learning frameworks. The improved performance is attributed to edge encodings of the interaction graph being an accurate intermediate representation of spatial behaviors between agents that are difficult to characterize in an end-end learning framework. The MRGCN is integrated with an attention layer that further improved the performance, often near-perfect performance. Significant performance gain on various datasets that are consistent across several metrics confirms the efficacy of the proposed framework. Seamless data transfer across datasets further showcases its reliability. Future directions include integrating the proposed framework with a behavior planner.

\nocite{*}

\end{document}